\documentclass[11pt,a4paper]{article}
\usepackage[]{acl2018}
\usepackage{times}
\usepackage{latexsym}
\usepackage{arydshln}
\usepackage{booktabs}
\usepackage{graphicx}
\usepackage{float}
\usepackage{color}
\usepackage{amsmath,amsfonts,amssymb,bbm,epsfig,bm}
\usepackage[ruled,noresetcount,linesnumbered]{algorithm2e}
\usepackage{listings}
\usepackage{subfig}
\usepackage{url}
\usepackage{spverbatim}
\usepackage{multirow}
\usepackage{pbox}
\usepackage{tikz}
\usepackage{pifont}
\usepackage{xspace}
\usepackage{subfig}
\usepackage{physics}
\usepackage[nameinlink,capitalize]{cleveref}
\usepackage{scalerel}
\usepackage[font=small]{caption}
\usepackage{titlesec}

\titlespacing{\paragraph}{%
  0pt}{
  0.3\baselineskip}{
  1em}

\usepackage{color}
\definecolor{deepblue}{rgb}{0,0,0.5}
\definecolor{deepred}{rgb}{0.6,0,0}
\definecolor{deepgreen}{rgb}{0,0.5,0}
\definecolor{darkgreen}{RGB}{43,163,39}
\definecolor{bluesquare}{rgb}{126,166,224}

\newcommand{\cmark}{\textcolor{darkgreen}{\ding{51}}}
\newcommand{\xmark}{\textcolor{red}{\ding{55}}}

\lstdefinestyle{pythoncode}{
	language=Python,
	otherkeywords={self,join,append,split,write},             
	keywordstyle=\bfseries\color{deepblue},
	emph={MyClass,__init__},          
	emphstyle=\color{deepred},    
	showstringspaces=false,
	breaklines=true,
	escapeinside=||,
	columns=fullflexible,
}

\lstdefinestyle{atiscode}{
  language=Python,
  otherkeywords={self,argmax,argmin,flight,from,to,departure,time,exists,round,trip,fare,has_meal,during,day,day,number,month,ground,transport,to,city,from,airport,meal,weekday,oneway,class_type,day,rental,car,ground,fare,has,meal},             
  keywordstyle=\bfseries\color{deepblue},
  emph={MyClass,__init__},          
  emphstyle=\color{deepred},    
  showstringspaces=false,
  breaklines=true,
  escapeinside=||,
  columns=fullflexible,
  basicstyle=\fontfamily{cmtt}\small,
  belowskip=-\baselineskip,
  aboveskip=-0.7\baselineskip
}

\usetikzlibrary{tikzmark}

\newcommand{\py}[1]{}

\newcommand\mr{\bm{z}}
\newcommand\x{\bm{x}}

\newcommand\applyconstrn{\textsc{ApplyConstr}}
\newcommand\DL{\mathbb{L}}
\newcommand\DU{\mathbb{U}}
\newcommand{\ie}{{\emph{i.e.}},\xspace}

\newcommand{\eg}{{\emph{e.g.}},\xspace}

\renewcommand{\tt}[1]{\fontfamily{cmtt}\selectfont #1}

\def\model/{\textsc{StructVAE}}
\def\atis/{\textsc{Atis}}
\def\django/{\textsc{Django}}
\def\sq/{{\sc Seq2Tree}}

\aclfinalcopy 
\def\aclpaperid{35} 

\title{\model/: Tree-structured Latent Variable Models for Semi-supervised Semantic Parsing}

\author{Pengcheng Yin,~\quad
  Chunting Zhou,\quad
  Junxian He,\quad
  Graham Neubig \\
  Language Technologies Institute \\
  Carnegie Mellon University \\
  \tt{\{pcyin,ctzhou,junxianh,gneubig\}@cs.cmu.edu}}

\date{}

\begin{document}
\maketitle
\begin{abstract}
  Semantic parsing is the task of transducing natural language (NL) utterances into formal meaning representations (MRs), commonly represented as tree structures.
  Annotating NL utterances with their corresponding MRs is expensive and time-consuming, and thus the limited availability of labeled data often becomes the bottleneck of data-driven, supervised models.
  We introduce \model/, a variational auto-encoding model for semi-supervised semantic parsing, which learns both from limited amounts of parallel data, and readily-available unlabeled NL utterances.
  \model/ models latent MRs not observed in the unlabeled data as tree-structured latent variables.
  Experiments on semantic parsing on the \atis/ domain and Python code generation show that with extra unlabeled data, \model/ outperforms strong supervised models.\footnote{Code available at \href{http://pcyin.me/struct_vae}{\tt http://pcyin.me/struct\_vae}}
\end{abstract}

\section{Introduction}
\label{sec:introduction}

Semantic parsing tackles the task of mapping natural language (NL) utterances into structured formal meaning representations (MRs).
This includes parsing to general-purpose logical forms such as $\lambda$-calculus~\cite{ZettlemoyerC05,DBLP:dblp_conf/emnlp/ZettlemoyerC07} and the abstract meaning representation~(AMR, \newcite{banarescu13amr,Misra:16neuralccg}), as well as parsing to computer-executable programs to solve problems such as question answering~\cite{berant2013freebase,DBLP:conf/acl/YihCHG15,DBLP:journals/corr/LiangBLFL16}, or generation of domain-specific (\eg~SQL) or general purpose programming languages (\eg~Python)~\citep{DBLP:conf/acl/QuirkMG15,yin17acl,rabinovich17syntaxnet}.
\begin{figure}[tb]
	\centering
	\includegraphics[width=0.85 \columnwidth]{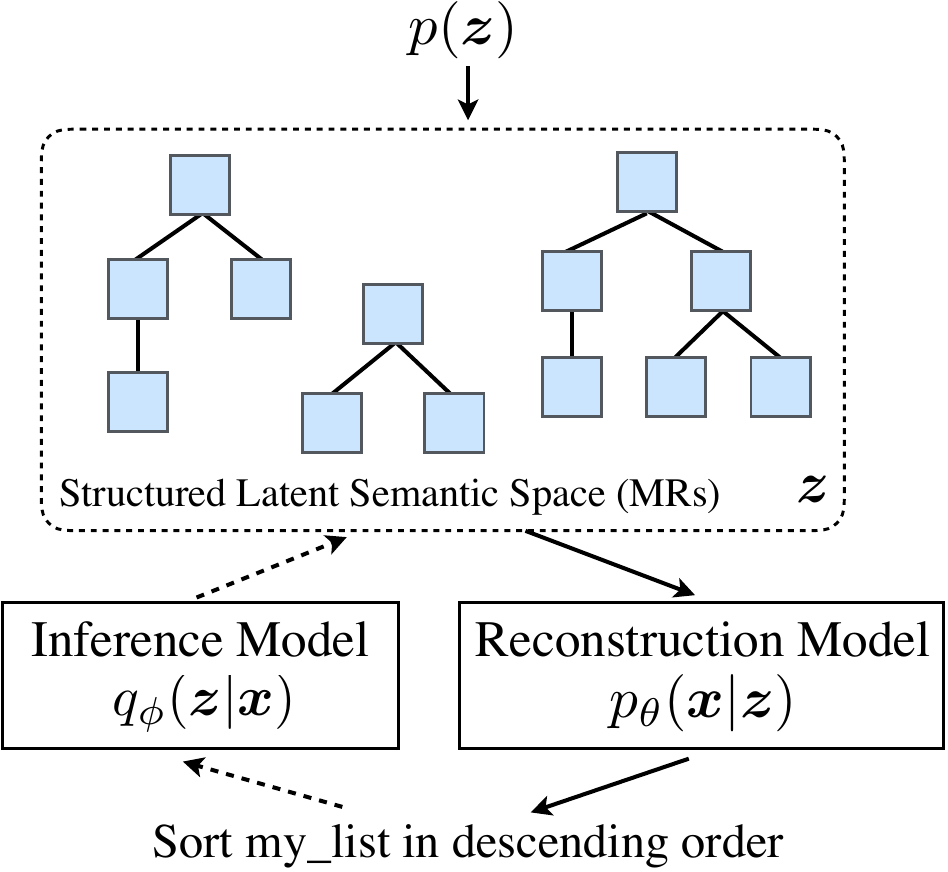}
	\caption{Graphical Representation of \model/}
	\label{fig:structVAE}
  \vspace{-5mm}
\end{figure}

While these models have a long history~\citep{DBLP:conf/aaai/ZelleM96,DBLP:dblp_conf/ecml/TangM01}, recent advances are largely attributed to the success of neural network models~\cite{DBLP:conf/acl/XiaoDG16,DBLP:conf/acl/LingBGHKWS16,DBLP:conf/acl/DongL16,iyer17user,DBLP:journals/corr/abs-1709-00103}.
However, these models are also extremely \emph{data hungry}: optimization of such models requires large amounts of training data of parallel NL utterances and manually annotated MRs, the creation of which can be expensive, cumbersome, and time-consuming.
Therefore, the limited availability of parallel data has become the bottleneck of existing, purely supervised-based models.
These data requirements can be alleviated with \emph{weakly-supervised} learning, where the denotations (\eg answers in question answering) of MRs (\eg logical form queries) are used as indirect supervision~(\newcite{DBLP:conf/conll/ClarkeGCR10,DBLP:conf/acl/LiangJK11,berant2013freebase}, {\it inter alia}), or \emph{data-augmentation techniques} that automatically generate pseudo-parallel corpora using hand-crafted or induced grammars~\cite{Jia2016,wang15overnight}. 


In this work, we focus on \emph{semi-supervised} learning, aiming to learn from both limited amounts of parallel NL-MR corpora, and \emph{unlabeled} but readily-available NL utterances.
We draw inspiration from recent success in applying variational auto-encoding (VAE) models in semi-supervised sequence-to-sequence learning~\cite{miao2016language,Kocisky2016}, and propose \model/ --- a principled deep generative approach for semi-supervised learning with tree-structured latent variables (\autoref{fig:structVAE}).
\model/ is based on a generative story where the surface NL utterances are generated from tree-structured latent MRs following the standard VAE architecture:
(1) an off-the-shelf semantic parser functions as the \emph{inference model}, parsing an observed NL utterance into latent meaning representations (\autoref{sec:structVAE:inference});
(2) a \emph{reconstruction model} decodes the latent MR into the original observed utterance (\autoref{sec:structVAE:reconstruction}).
This formulation enables our model to perform both standard supervised learning by optimizing the inference model (\ie~the parser) using parallel corpora, and unsupervised learning by maximizing the variational lower bound of the likelihood of the unlabeled utterances (\autoref{sec:structVAE:learning}).


In addition to these contributions to semi-supervised semantic parsing, \model/ contributes to generative model research as a whole, providing a recipe for training VAEs with \emph{structured} latent variables.
Such a structural latent space is contrast to existing VAE research using \emph{flat} representations, such as continuous distributed representations~\cite{kingma2013auto}, discrete symbols~\cite{miao2016language}, or hybrids of the two~\cite{zhou17acl}.

We apply \model/ to semantic parsing on the \atis/ domain and Python code generation.
As an auxiliary contribution, we implement a transition-based semantic parser, 
which uses Abstract Syntax Trees (ASTs,~\autoref{sec:inference}) as intermediate MRs and achieves strong results on the two tasks.
We then apply this parser as the inference model for semi-supervised learning, and show that with extra unlabeled data, \model/ outperforms its supervised counterpart.
We also demonstrate that \model/ is compatible with different structured latent representations, applying it to a simple sequence-to-sequence parser which uses $\lambda$-calculus logical forms as MRs.



\vspace{-1mm}
\section{Semi-supervised Semantic Parsing}
\vspace{-1mm}
\label{sec:semi_sup_learning}
In this section we introduce the objectives for semi-supervised semantic parsing, and present high-level intuition in applying VAEs for this task.

\subsection{Supervised and Semi-supervised Training}
\label{sec:semi_sup_learning:sup_and_semi_sup}

Formally, semantic parsing is the task of mapping utterance $\x$ to a meaning representation $\mr$.
As noted above, there are many varieties of MRs that can be represented as either graph structures (\eg~AMR) or tree structures (\eg~$\lambda$-calculus and ASTs for programming languages).
In this work we specifically focus on tree-structured MRs (see \autoref{fig:ast_gen_example} for a running example Python AST), although application of a similar framework to graph-structured representations is also feasible.




Traditionally, purely supervised semantic parsers train a probabilistic model $p_\phi(\mr|\x)$ using parallel data $\DL$ of NL utterances and annotated MRs (\ie $\DL=\{\langle \x, \mr \rangle\}$).
As noted in the introduction, one major bottleneck in this approach is the lack of such parallel data.
Hence, we turn to semi-supervised learning, where the model additionally has access to a relatively large amount of unlabeled NL utterances $\DU = \{ \x \}$. Semi-supervised learning then aims to maximize the log-likelihood of examples in both $\DL$ and $\DU$:
\begin{equation}
	\mathcal{J} = \underbrace{\sum_{\langle \x, \mr \rangle\ \in \DL} \log p_\phi(\mr|\x)}_{\textrm{supervised obj.}~~\mathcal{J}_s} 
	+ \alpha \hspace{-3mm} \underbrace{\sum_{\x \in \DU} \log p(\x)}_{\textrm{unsupervised obj.}~~\mathcal{J}_u}
	\label{eq:semisup:obj}
\end{equation}
The joint objective consists of two terms: (1) a supervised objective $\mathcal{J}_s$ that maximizes the conditional likelihood of annotated MRs, as in standard supervised training of semantic parsers; and (2) a unsupervised objective $\mathcal{J}_u$, which maximizes the marginal likelihood $p(\x)$ of unlabeled NL utterances $\DU$, controlled by a tuning parameter $\alpha$.
Intuitively, if the modeling of $p_\phi(\mr|\x)$ and $p(\x)$ is coupled (\eg they share parameters), then optimizing the marginal likelihood $p(\x)$ using the unsupervised objective $\mathcal{J}_u$ would help the learning of the semantic parser $p_\phi(\mr|\x)$~\citep{zhu05survey}.
\model/ uses the variational auto-encoding framework to jointly optimize $p_\phi(\mr|\x)$ and $p(\x)$, as outlined in~\autoref{sec:semi_sup_learning:vae} and detailed in~\autoref{sec:structVAE}.

\subsection{VAEs for Semi-supervised Learning}
\label{sec:semi_sup_learning:vae}
From~\cref{eq:semisup:obj}, our semi-supervised model must be able to calculate the probability $p(\x)$ of unlabeled NL utterances.
To model $p(\x)$, we use VAEs, which provide a principled framework for generative models using neural networks~\citep{kingma2013auto}.
As shown in~\autoref{fig:structVAE}, VAEs define a \emph{generative story} (bold arrows in~\autoref{fig:structVAE}, explained in \autoref{sec:structVAE:reconstruction}) to model $p(\x)$, where a latent MR $\mr$ is sampled from a prior, and then passed to the \emph{reconstruction} model to decode into the surface utterance $\x$.
There is also an \emph{inference} model $q_{\phi}(\mr|\x)$ that allows us to infer the most probable latent MR $\mr$ given the input $\x$ (dashed arrows in~\autoref{fig:structVAE}, explained in \autoref{sec:structVAE:inference}).
In our case, the inference process is equivalent to the task of semantic parsing if we set $q_{\phi}(\cdot)\triangleq p_{\phi}(\cdot)$.
VAEs also provide a framework to compute an approximation of $p(\x)$ using the inference and reconstruction models, allowing us to effectively optimize the unsupervised and supervised objectives in \cref{eq:semisup:obj} in a joint fashion~(\newcite{kingma2014semi}, explained in~\autoref{sec:structVAE:learning}).

%
%

\vspace{-0.2em}
\section{\model/: VAEs with Tree-structured Latent Variables}
\label{sec:structVAE}
\vspace{-0.2em}

\subsection{Generative Story}
\label{sec:structVAE:reconstruction}

\model/ follows the standard VAE architecture, and defines a generative story that explains how an NL utterance is generated: a latent meaning representation $\mr$ is sampled from a prior distribution $p(\mr)$ over MRs, which encodes the latent semantics of the utterance. 
A \emph{reconstruction} model $p_\theta(\x|\mr)$ then decodes the sampled MR $\mr$ into the observed NL utterance $\x$.

Both the prior $p(\mr)$ and the reconstruction model $p(\x|\mr)$ takes tree-structured MRs as inputs.
To model such inputs with rich internal structures, we follow~\newcite{konstas2017neural}, and model the distribution over a sequential surface representation of $\mr$, $\mr^s$ instead.
Specifically, we have $p(\mr) \triangleq p(\mr^s)$ and $p_\theta(\x|\mr) \triangleq p_\theta(\x|\mr^s)$\footnote{Linearizion is used by the prior and the reconstruction model only, and not by the inference model.}.
For code generation, $\mr^s$ is simply the surface source code of the AST $\mr$.
For semantic parsing, $\mr^s$ is the linearized s-expression of the logical form.
Linearization allows us to use standard sequence-to-sequence networks to model $p(\mr)$ and $p_\theta(\x|\mr)$.
As we will explain in~\autoref{sec:exp:results}, we find these two components perform well with linearization.

Specifically, the prior is parameterized by a Long Short-Term Memory (LSTM) language model over $\mr^s$.
The reconstruction model is an attentional sequence-to-sequence network~\citep{luong2015effective}, augmented with a copying mechanism~\citep{DBLP:conf/acl/GuLLL16}, allowing an out-of-vocabulary (OOV) entity in $\mr^s$ to be copied to $\x$ (\eg the variable name {\tt my\_list} in~\autoref{fig:structVAE} and its AST in~\autoref{fig:ast_gen_example}). 
We refer readers to~\autoref{sec:app:neuralnet} for details of the neural network architecture.

\subsection{Inference Model}
\label{sec:structVAE:inference}
\begin{figure*}[t]
\begin{minipage}[t]{0.95 \columnwidth}
	\centering
	\vspace{0pt}
	\setlength{\tabcolsep}{0.3em} 
	\includegraphics[width=\columnwidth]{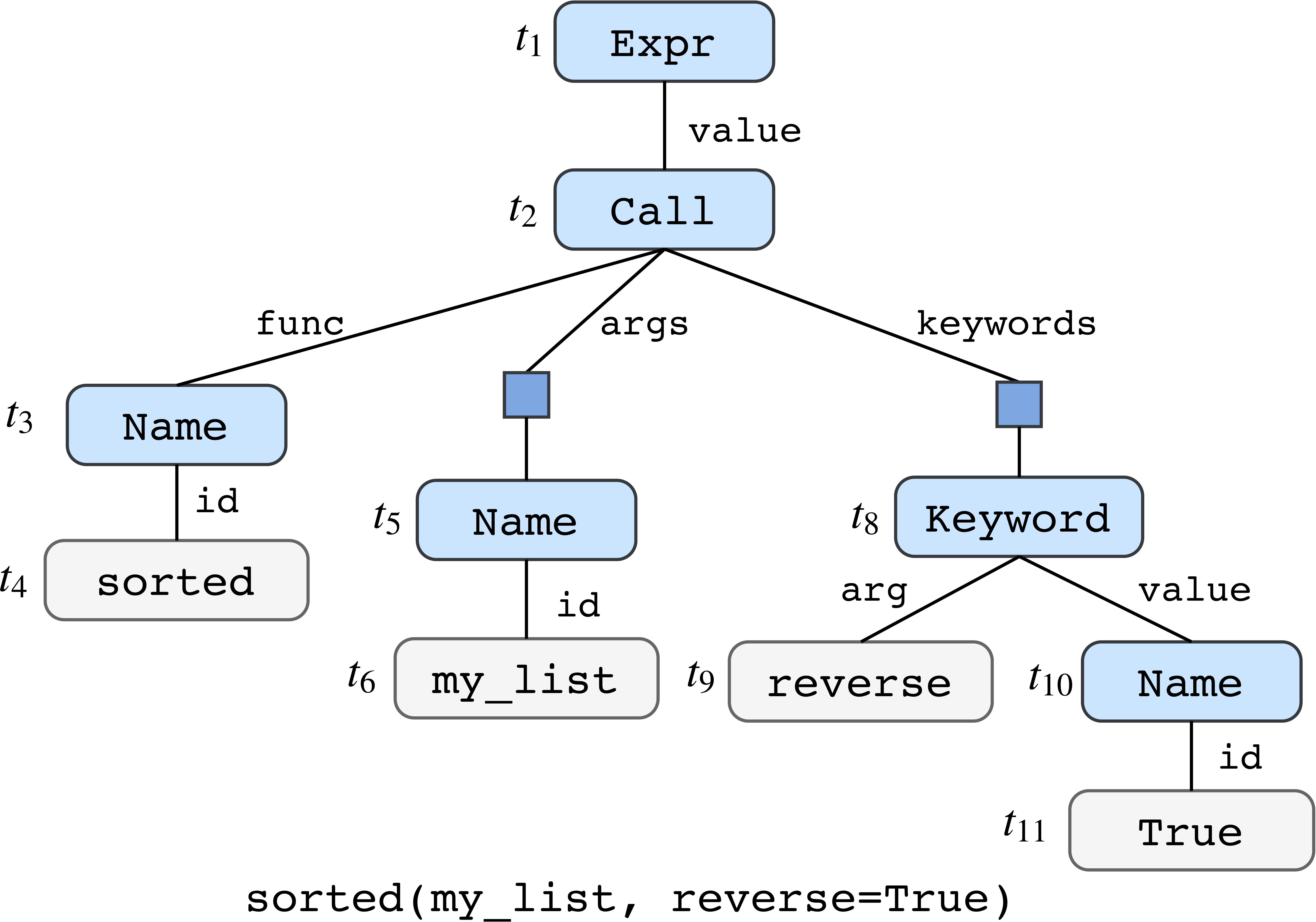}
\end{minipage}%
\hfill%
\begin{minipage}[t]{1.05 \columnwidth}
	\centering
	{\renewcommand{\arraystretch}{0.9} 
	\vspace{0pt}
	\resizebox{\columnwidth}{!}{%
	\begin{tabular}{lll}
	\hline

	\hline
	$\bm{t}$ & {\bf Frontier Field} & {\bf Action} \\
	\hdashline
	$t_1$ & {\tt stmt root}     & {\tt Expr(expr value)}  \\
	$t_2$ & {\tt expr value}    & {\tt Call(expr func, expr* args, }  \\
	    &						& \hfill {\tt keyword* keywords)} \\
	$t_3$ & {\tt expr func}     & {\tt Name(identifier id)} \\
	$t_4$ & {\tt identifier id} & {\sc GenToken}$[sorted]$  \\
	$t_5$ & {\tt expr* args} & {\tt Name(identifier id)}  \\
	$t_6$ & {\tt identifier id} & {\sc GenToken}$[my\_list]$  \\
	$t_7$ & {\tt expr* args} & {\sc Reduce} (close the frontier field)  \\
	$t_8$ & {\tt keyword* keywords} & {\tt keyword(identifier arg,}  \\
	      &                         & \hfill {\tt expr value)} \\
	$t_9$ & {\tt identifier arg} & {\sc GenToken}$[reverse]$  \\
	$t_{10}$ & {\tt expr value} & {\tt Name(identifier id)}  \\
	$t_{11}$ & {\tt identifier id} & {\sc GenToken}$[True]$ \\
	$t_{12}$ & {\tt keyword* keywords} & {\sc Reduce} (close the frontier field) \\
	\hline

	\hline
	\end{tabular}}}
\end{minipage}
\caption{\textbf{\textit{Left}} An example ASDL AST with its surface source code. Field names are labeled on upper arcs. Blue squares denote fields with {\it sequential} cardinality. Grey nodes denote primitive identifier fields, with annotated values. Fields are labeled with time steps at which they are generated. \textbf{\textit{Right}} Action sequences used to construct the example AST. Frontier fields are denoted by their signature {\tt (type name)}. Each constructor in the Action column refers to an {\sc ApplyConstr} action.}
\vspace{-3mm}
\label{fig:ast_gen_example}
\end{figure*}

\model/ models the semantic parser $p_\phi(\mr|\x)$ as the inference model $q_\phi(\mr|\x)$ in VAE (\autoref{sec:semi_sup_learning:vae}), which maps NL utterances $\x$ into tree-structured meaning representations $\mr$.
$q_\phi(\mr|\x)$ can be any trainable semantic parser, with the corresponding MRs forming the structured latent semantic space.
In this work, we primarily use a semantic parser based on the Abstract Syntax Description Language (ASDL) framework~\citep{wang97asdl} as the inference model.
The parser encodes $\x$ into ASTs (\autoref{fig:ast_gen_example}). 
ASTs are the native meaning representation scheme of source code in modern programming languages, and can also be adapted to represent other semantic structures, 
like $\lambda$-calculus logical forms (see \autoref{sec:exp:setup} for details).
We remark that \model/ works with other semantic parsers with different meaning representations as well (\eg~using $\lambda$-calculus logical forms for semantic parsing on \atis/, explained in~\autoref{sec:exp:results}).

\label{sec:inference}

Our inference model is a transition-based parser inspired by recent work in neural semantic parsing and code generation.
The transition system is an adaptation of~\citet{yin17acl} (hereafter YN17), which decomposes the generation process of an AST into sequential applications of tree-construction actions following the ASDL grammar, thus ensuring the syntactic well-formedness of generated ASTs.
Different from YN17, where ASTs are represented as a Context Free Grammar learned from a parsed corpus, we follow~\citet{rabinovich17syntaxnet} and use ASTs defined under the ASDL formalism (\autoref{sec:structVAE:inference:asdl}).

\subsubsection{Generating ASTs with ASDL Grammar}
\label{sec:structVAE:inference:asdl}
\begin{figure}[t]
	\centering
	\includegraphics[width=\columnwidth]{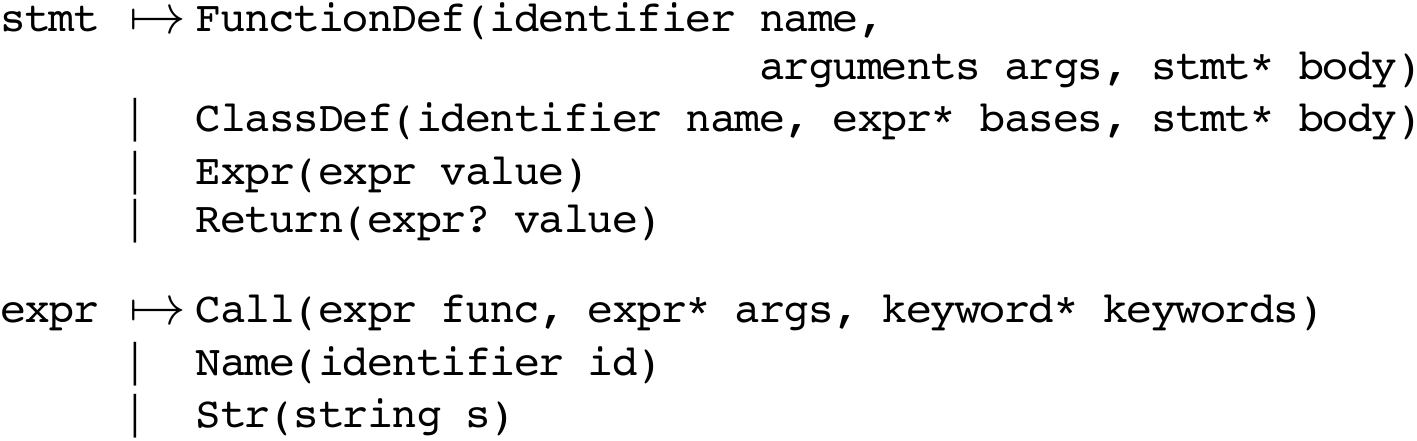}
	\caption{Excerpt of the python abstract syntax grammar~\citep{pythonast}}
	\label{fig:py_asdl}
  \vspace{-4mm}
\end{figure}

First, we present a brief introduction to ASDL.
An AST can be generated by applying typed {\em constructors} in an ASDL grammar, such as those in \autoref{fig:py_asdl} for the Python ASDL grammar.
Each constructor specifies a language construct, and is assigned to a particular {\em composite type}.
For example, the constructor {\tt Call} has type {\tt expr} (expression), and it denotes function calls.
Constructors are associated with multiple {\em fields}.
For instance, the {\tt Call} constructor and has three fields: {\it func}, {\it args} and {\it keywords}.
Like constructors, fields are also strongly typed.
For example, the {\it func} field of {\tt Call} has {\tt expr} type.
Fields with composite types are instantiated by constructors of the same type, while fields with {\it primitive} types store values (\eg identifier names or string literals).
Each field also has a cardinality (single, optional $?$, and sequential $*$), specifying the number of values the field has.

Each node in an AST corresponds to a typed field in a constructor (except for the root node). 
Depending on the cardinality of the field, an AST node can be instantiated with one or multiple constructors.
For instance, the {\it func} field in the example AST has single cardinality, and is instantiated with a {\tt Name} constructor; while the {\it args} field with sequential cardinality could have multiple constructors (only one shown in this example).


%

Our parser employs a transition system to generate an AST using three types of actions.~\autoref{fig:ast_gen_example} (Right) lists the sequence of actions used to generate the example AST.
The generation process starts from an initial derivation with only a root node of type {\tt stmt} (statement), and proceeds according to the top-down, left-to-right traversal of the AST. At each time step, the parser applies an action to the {\em frontier field} of the derivation:


\textbf{\textsc{ApplyConstr}$[c]$} actions apply a constructor $c$ to the frontier composite field, expanding the derivation using the fields of $c$.
For fields with single or optional cardinality, an \applyconstrn~action instantiates the empty frontier field using the constructor, while for fields with sequential cardinality, it appends the constructor to the frontier field.
For example, at $t_2$ the {\tt Call} constructor is applied to the {\it value} field of {\tt Expr}, and the derivation is expanded using its three child fields.

\textbf{\textsc{Reduce}} actions complete generation of a field with optional or multiple cardinalities. For instance, the {\it args} field is instantiated by {\tt Name}
at $t_5$, and then closed by a {\sc Reduce} action at $t_7$.

\textbf{\textsc{GenToken}}$[v]$ actions populate an empty primitive frontier field with token $v$.
A primitive field whose value is a single token (\eg identifier fields) can be populated with a single \textsc{GenToken} action.
Fields of {\tt string} type can be instantiated using multiple such actions, with a final $\textsc{GenToken}[${\tt </f>}$]$ action to terminate the generation of field values.

\subsubsection{Modeling $q_{\phi}(\mr|\x)$}
The probability of generating an AST $\mr$ is naturally decomposed into the probabilities of the actions $\{ a_t \}$ used to construct $\mr$:
\begin{equation*}\setlength\abovedisplayskip{4pt}\setlength\belowdisplayskip{4pt}
	q_{\phi}(\mr|\x) = \prod_{t} p(a_t | a_{<t}, \x).
\end{equation*}
Following YN17, we parameterize $q_{\phi}(\mr|\x)$ using a sequence-to-sequence network with auxiliary recurrent connections following the topology of the AST.
Interested readers are referred to~\autoref{sec:app:neuralnet} and \citet{yin17acl} for details of the neural network architecture.


\subsection{Semi-supervised Learning}
\label{sec:structVAE:learning}
In this section we explain how to optimize the semi-supervised learning objective~\cref{eq:semisup:obj} in \model/.

\paragraph{Supervised Learning} For the supervised learning objective, we modify $\mathcal{J}_s$, and use the labeled data to optimize both the inference model (the semantic parser) and the reconstruction model:
\begin{equation}\setlength\abovedisplayskip{4pt}\setlength\belowdisplayskip{4pt}
	\mathcal{J}_s \triangleq \sum_{(\x, \mr) \in \DL} \big( \log q_{\phi}(\mr|\x) + \log p_{\theta}(\x|\mr) \big)
  \label{eq:structVAE:sup_obj}
\end{equation}

\paragraph{Unsupervised Learning}
To optimize the unsupervised learning objective $\mathcal{J}_u$ in \cref{eq:semisup:obj}, we maximize the variational lower-bound of $\log p(\x)$:
\begingroup
\setlength\abovedisplayskip{4pt}\setlength\belowdisplayskip{4pt}
\begin{multline}
	\log p(\x) \ge \mathbb{E}_{\mr \sim q_{\phi}(\mr|\x)} \big( \log p_{\theta}(\x|\mr) \big) \\ - \lambda \cdot \mathrm{KL}[q_{\phi}(\mr|\x) || p(\mr)] = \mathcal{L}
	\label{eq:structVAE:va_lower_bound}
\end{multline}
\endgroup
where $\mathrm{KL}[q_\phi || p]$ is the Kullback-Leibler (KL) divergence.
Following common practice in optimizing VAEs, we introduce $\lambda$ as a tuning parameter of the KL divergence to control the impact of the prior~\citep{miao2016language,DBLP:conf/conll/BowmanVVDJB16}.

To optimize the parameters of our model in the face of non-differentiable discrete latent variables, we follow~\citet{miao2016language}, and approximate $\pdv{\mathcal{L}}{\phi}$ using the score function estimator (a.k.a.~REINFORCE, \citet{DBLP:journals/ml/Williams92}):
\begin{equation}
\setlength\abovedisplayskip{5pt}\setlength\belowdisplayskip{5pt}
\begin{split}
 	\pdv{\mathcal{L}}{\phi} &= \pdv{}{\phi} \mathbb{E}_{\mr \sim q_{\phi}(\mr|\x)} \\ 
  &\quad \underbrace{\Big(\log p_{\theta}(\x|\mr) - \lambda \big( \log q_{\phi}(\mr|\x) - \log p(\mr) \big) \Big)}_\textrm{learning signal} \\
 	&= \pdv{}{\phi} \mathbb{E}_{\mr \sim q_{\phi}(\mr|\x)} l'(\x, \mr) \\
 	&\approx \frac{1}{|\mathcal{S}(\x)|} \hspace{-1mm} \sum_{\mr_i \in \mathcal{S}(\x)} \hspace{-2mm} l'(\x, \mr_i) \pdv{\log q_{\phi}(\mr_i|\x)}{\phi}
 	\raisetag{2.1em}\label{eq:framework:semi_sup:encoder_grad}
\end{split}
\end{equation}
where we approximate the gradient using a set of samples $\mathcal{S}(\x)$ drawn from $q_{\phi}(\cdot|\x)$. 
To ensure the quality of sampled latent MRs, we follow~\citet{DBLP:conf/acl/GuuPLL17} and use beam search.
The term $l'(\x, \mr)$ is defined as the {\em learning signal}~\citep{miao2016language}.
The learning signal weights the gradient for each latent sample $\mr$.
In REINFORCE, to cope with the high variance of the learning signal, it is common to use a baseline $b(\x)$ to stabilize learning, and re-define the learning signal as 
\begin{equation}\setlength\abovedisplayskip{4pt}\setlength\belowdisplayskip{4pt}
l(\x, \mr) \triangleq l'(\x, \mr) - b(\x).
\label{eq:structVAE:lr}
\end{equation}
Specifically, in \model/, we define
\begin{equation}\setlength\abovedisplayskip{4pt}\setlength\belowdisplayskip{4pt}
	b(\x) = a \cdot \log p(\x) + c,
  \label{eq:structVAE:baseline}
\end{equation}
where $\log p(\x)$ is a pre-trained LSTM language model.
This is motivated by the empirical observation that $\log p(\x)$ correlates well with the reconstruction score $\log p_{\theta}(\x|\mr)$, hence with $l'(\x, \mr)$. 

Finally, for the reconstruction model, its gradient can be easily computed:
\begin{equation*}\setlength\abovedisplayskip{4pt}\setlength\belowdisplayskip{4pt}
	\pdv{\mathcal{L}}{\theta} \approx \frac{1}{|\mathcal{S}(\x)|} \sum_{\mr_i \in \mathcal{S}(\x)} \pdv{\log p_{\theta}(\x|\mr_i)}{\theta}.
\end{equation*}

\paragraph{Discussion}
Perhaps the most intriguing question here is why semi-supervised learning could improve semantic parsing performance. 
While the underlying theoretical exposition still remains an active research problem~\cite{singh2008unlabeled}, in this paper we try to empirically test some likely hypotheses.
In~\cref{eq:framework:semi_sup:encoder_grad}, the gradient received by the inference model from each latent sample $\mr$ is weighed by the learning signal $l(\x, \mr)$.
$l(\x, \mr)$ can be viewed as the reward function in REINFORCE learning.
It can also be viewed as weights associated with pseudo-training examples $\{ \langle \x, \mr \rangle : \mr \in \mathcal{S}(\x) \}$ sampled from the inference model.
Intuitively, a sample $\mr$ with higher rewards should:
(1) have $\mr$ adequately encode the input, leading to high reconstruction score $\log p_{\theta}(\x|\mr)$; and (2) have $\mr$ be succinct and natural, yielding high prior probability.
Let $\mr^*$ denote the gold-standard MR of $\x$. Consider the ideal case where $\mr^* \in \mathcal{S}(\x)$ and $l(\x, \mr^*)$ is positive, while $l(\x, \mr')$ is negative for other imperfect samples $\mr' \in \mathcal{S}(\x)$, $\mr' \neq \mr^*$.
In this ideal case, $\langle \x, \mr^* \rangle$ would serve as a positive training example and other samples $\langle \x, \mr' \rangle$ would be treated as negative examples. 
Therefore, the inference model would receive informative gradient updates, and learn to discriminate between gold and imperfect MRs.
This intuition is similar in spirit to recent efforts in interpreting gradient update rules in reinforcement learning~\citep{DBLP:conf/acl/GuuPLL17}.
We will present more empirical statistics and observations in \autoref{sec:exp:results}.


\vspace{-1mm}
\section{Experiments}
\vspace{-1mm}
\label{sec:exp}



\subsection{Datasets}
\label{sec:exp:dataset_metric}

In our semi-supervised semantic parsing experiments, it is of interest how \model/ could 
further improve upon a supervised parser with extra unlabeled data. 
We evaluate on two datasets:

\paragraph{Semantic Parsing} We use the \atis/ dataset, a collection of 5,410 telephone inquiries of flight booking (\eg~{\it ``Show me flights from ci0 to ci1''}).
The target MRs are defined using $\lambda$-calculus logical forms (\eg~``{\tt lambda \$0 e (and (flight \$0) (from \$ci0) (to \$ci1))}'').
We use the pre-processed dataset released by~\citet{DBLP:conf/acl/DongL16}, where entities (\eg~cities) are canonicalized using typed slots (\eg~{\tt ci0}).
To predict $\lambda$-calculus logical forms using our transition-based parser, we use the ASDL grammar defined by~\citet{rabinovich17syntaxnet} to convert between logical forms and ASTs (see~\autoref{sec:app:asdl_atis} for details).

\paragraph{Code Generation} The \django/ dataset~\cite{DBLP:conf/kbse/OdaFNHSTN15} contains 18,805 lines of Python source code extracted from the Django web framework.
Each line of code is annotated with an NL utterance.
Source code in the \django/ dataset exhibits a wide variety of real-world use cases of Python, including IO operation, data structure manipulation, class/function definition, \textit{etc}.
We use the pre-processed version released by~\citet{yin17acl} and use the {\tt astor} package to convert ASDL ASTs into Python source code.

\subsection{Setup}
\label{sec:exp:setup}


\paragraph{Labeled and Unlabeled Data}
\model/ requires access to extra unlabeled NL utterances for semi-supervised learning. 
However, the datasets we use do not accompany with such data.
We therefore simulate the semi-supervised learning scenario by randomly sub-sampling $K$ examples from the training split of each dataset as the labeled set $\DL$.
To make the most use of the NL utterances in the dataset, we construct the unlabeled set $\DU$ using all NL utterances in the training set\footnote{We also tried constructing $\DU$ using the disjoint portion of the NL utterances not presented in the labeled set $\DL$, but found this yields slightly worse performance, probably due to lacking enough unlabeled data. Interpreting these results would be an interesting avenue for future work.}$^{,}$\footnote{While it might be relatively easy to acquire additional unlabeled utterances in practical settings (\eg through query logs of a search engine), unfortunately most academic semantic parsing datasets, like the ones used in this work, do not feature large sets of in-domain unlabeled data. We therefore perform simulated experiments instead.}.

\paragraph{Training Procedure} 
Optimizing the unsupervised learning objective~\cref{eq:structVAE:va_lower_bound} requires sampling structured MRs from the inference model $q_{\phi}(\mr|\x)$.
Due to the complexity of the semantic parsing problem, we cannot expect any valid samples from randomly initialized $q_{\phi}(\mr|\x)$.
We therefore pre-train the inference and reconstruction models using the supervised objective~\cref{eq:structVAE:sup_obj} until convergence, and then optimize using the semi-supervised learning objective~\cref{eq:semisup:obj}. 
Throughout all experiments we set $\alpha$ (\cref{eq:semisup:obj}) and $\lambda$ (\cref{eq:structVAE:va_lower_bound}) to 0.1.
The sample size $|\mathcal{S}(\x)|$ is 5.
We observe that the variance of the learning signal could still be high when low-quality samples are drawn from the inference model $q_{\phi}(\mr|\x)$.
We therefore clip all learning signals lower than $k=-20.0$.
Early-stopping is used to avoid over-fitting.
We also pre-train the prior $p(\mr)$~(\autoref{sec:structVAE:learning}) and the baseline function~\cref{eq:structVAE:baseline}.
Readers are referred to~\autoref{sec:app:config} for more detail of the configurations.

\paragraph{Metric} As standard in semantic parsing research, we evaluate by exact-match {\bf accuracy}. 

\vspace{-0.5mm}
\subsection{Main Results}
\vspace{-0.5mm}
\label{sec:exp:results}

\begin{table}[tb]
	\centering
	\resizebox{0.95 \columnwidth}{!}{
	\begin{tabular}{lccc}
	\toprule
	\textbf{$|\DL|$} & \textsc{Sup.} & \textsc{SelfTrain} & \model/ \\
	\hline
		500 & 63.2   & 65.3 & {\bf 66.0} \\
		1,000 & 74.6 & 74.2 & {\bf 75.7} \\
		2,000 & 80.4 & {\bf 83.3} & 82.4 \\
		3,000 & 82.8 & {\bf 83.6} & {\bf 83.6} \\ \hdashline
		4,434 (All) & {\bf 85.3} & -- & 84.5 \\
	\midrule
	\multicolumn{4}{@{}l@{}}{
	\begin{tabular}{lc}
		\textbf{Previous Methods} & \textsc{Acc.} \\
		  ZC07~\citep{DBLP:dblp_conf/emnlp/ZettlemoyerC07} & 84.6 \\
		  WKZ14~\citep{wang-kwiatkowski-zettlemoyer:2014:EMNLP2014} & {\bf 91.3} \\
		  \sq/~\citep{DBLP:conf/acl/DongL16}$^\dagger$ & 84.6 \\ 
		  ASN~\citep{rabinovich17syntaxnet}$^\dagger$ & 85.3 \\
		  ~~~~~~~~+ supervised attention & 85.9 \\
	\end{tabular}
	} 
	\\ \bottomrule 
	\end{tabular}}
	\caption{Performance on \atis/ w.r.t.~the size of labeled training data $\DL$. $^\dagger$Existing neural network-based methods}
	\label{tab:exp:results:atis}
\end{table}
\begin{table}[tb]
	\centering
  \small
	\resizebox{0.95 \columnwidth}{!}{
	\begin{tabular}{lccc}
	\toprule
	\textbf{$|\DL|$} & \textsc{Sup.} & \textsc{SelfTrain} & {\model/} \\
	\hline
		
    1,000 & 49.9 & 49.5 & {\bf 52.0} \\
		2,000 & 56.6 & 55.8 & {\bf 59.0} \\
		3,000 & 61.0 & 61.4 & {\bf 62.4} \\
		5,000 & 63.2 & 64.5 & {\bf 65.6} \\
		8,000 & 70.3 & 69.6 & {\bf 71.5} \\
		12,000 & 71.1 & 71.6  &	{\bf 72.0} \\ \hdashline
		16,000 (All) & {\bf 73.7} & -- & 72.3 \\

	\midrule	
	\multicolumn{4}{@{}l@{}}{
	\begin{tabular}{lc}
		\textbf{Previous Method} & \textsc{Acc.} \\
		YN17~\citep{yin17acl}~~~~~~~~~~~~~~~ & 71.6 \\
	\end{tabular}
	} \\ \bottomrule
	\end{tabular}}
	\caption{Performance on \django/ w.r.t.~the size of labeled training data $\DL$}
	\label{tab:exp:results:django}
  \vspace{-3mm}
\end{table}

\autoref{tab:exp:results:atis} and \autoref{tab:exp:results:django} list the results on \atis/ and \django/, resp, with varying amounts of labeled data $\DL$. 
We also present results of training the transition-based parser using only the supervised objective (\textbf{\textsc{Sup.}},~\cref{eq:structVAE:sup_obj}).
We also compare \model/ with self-training (\textbf{\textsc{SelfTrain}}), a semi-supervised learning baseline which uses the supervised parser to predict MRs for unlabeled utterances in $\DU - \DL$, and adds the predicted examples to the training set to fine-tune the supervised model.
Results for \model/ are averaged over four runs to account for the additional fluctuation caused by REINFORCE training.



\paragraph{Supervised System Comparison} 
First, to highlight the effectiveness of our transition parser based on ASDL grammar (hence the reliability of our supervised baseline), we compare the supervised version of our parser with existing parsing models.
On \atis/, our supervised parser trained on the full data is competitive with existing neural network based models, surpassing the \sq/ model, and on par with the Abstract Syntax Network (ASN) without using extra supervision.
On \django/, our model significantly outperforms the YN17 system, probably because the transition system used by our parser is defined natively to construct ASDL ASTs, reducing the number of actions for generating each example.
On \django/, the average number of actions is 14.3, compared with 20.3 reported in YN17.

\begin{figure}[t]
  \captionsetup{farskip=0pt}%
  \centering
    \subfloat[\django/]{\label{fig:exp:lr:hist:django}{\includegraphics[width=0.8 \columnwidth]{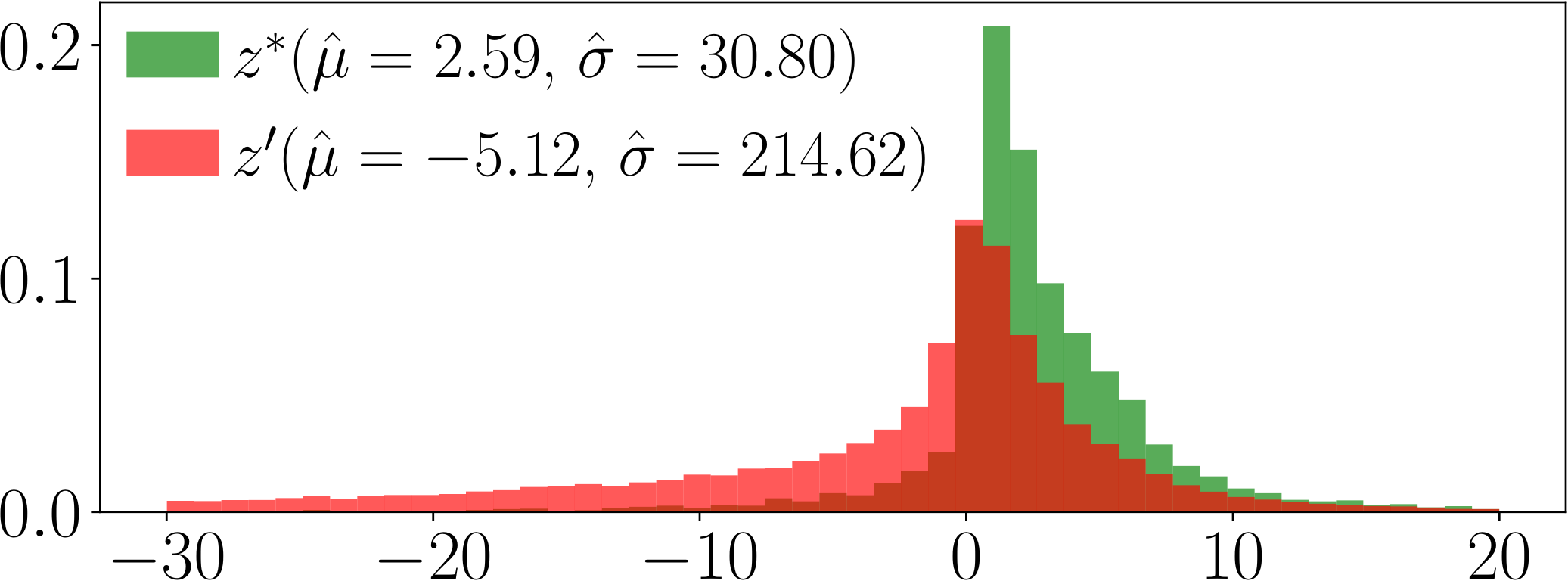} }}%
    \hfill
    \subfloat[\atis/]{\label{fig:exp:lr:hist:atis}{\includegraphics[width=0.8 \columnwidth]{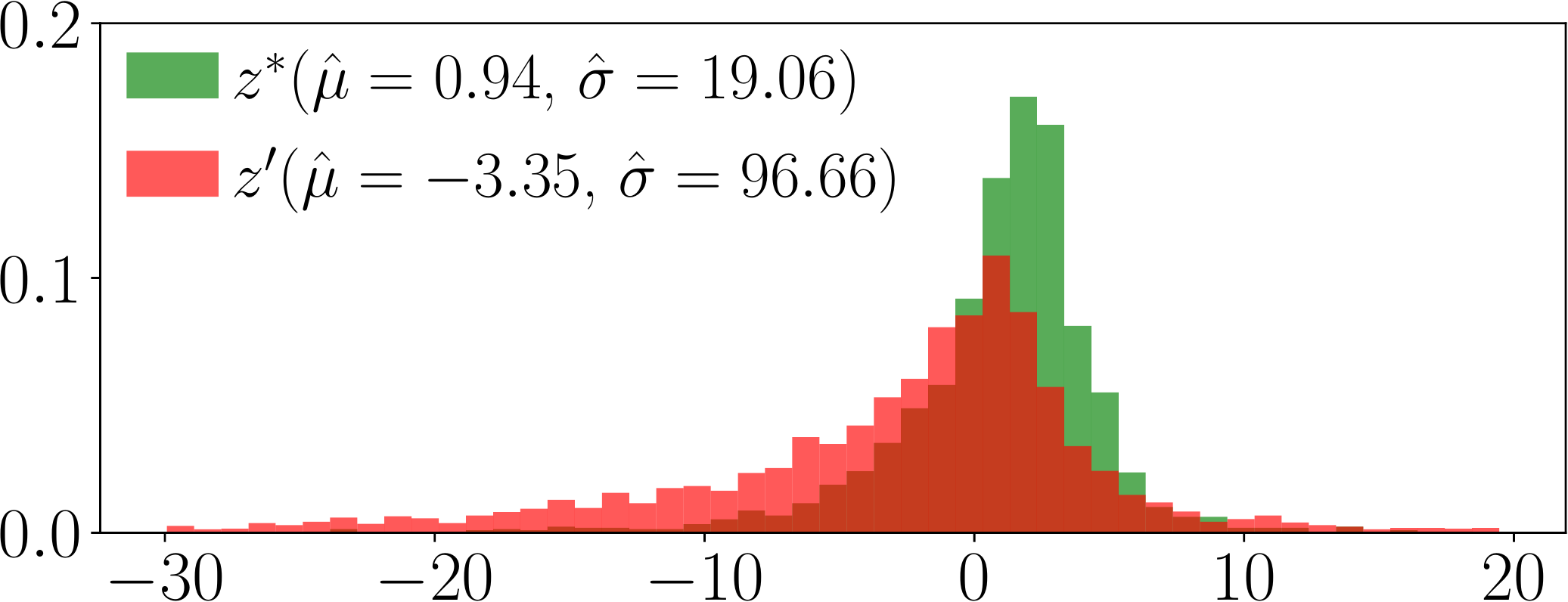} }}%
    \vspace{-2mm}%
    \caption{Histograms of learning signals on \django/ ($|\DL|=5000$) and \atis/ ($|\DL|=2000$). Difference in sample means is statistically significant ($p < 0.05$).}
  \label{fig:exp:lr:hist}
  \vspace{-4mm}
\end{figure}

\paragraph{Semi-supervised Learning} 
Next, we discuss our main comparison between \model/ with the supervised version of the parser (recall that the supervised parser is used as the inference model in \model/,~\autoref{sec:structVAE:inference}).
First, comparing our proposed \model/ with the supervised parser when there are extra unlabeled data (\ie~$|\DL| < 4,434$ for \atis/ and $|\DL| < 16,000$ for \django/), semi-supervised learning with \model/ consistently achieves better performance.
Notably, on \django/, our model registers results as competitive as previous state-of-the-art method (YN17) using only {\em half} the training data ($71.5$ when $|\DL|=8000$~v.s.~71.6 for YN17).
This demonstrates that \model/ is capable of learning from unlabeled NL utterances by inferring high quality, structurally rich latent meaning representations, further improving the performance of its supervised counterpart that is already competitive.
Second, comparing \model/ with self-training, we find \model/ outperforms \textsc{SelfTrain} in eight out of ten settings, while \textsc{SelfTrain} under-performs the supervised parser in four out of ten settings. This shows self-training does not necessarily yield stable gains while \model/ does. 
Intuitively, \model/ would perform better since it benefits from the additional signal of the quality of MRs from the reconstruction model (\autoref{sec:structVAE:learning}), for which we present more analysis in our next set of experiments.

For the sake of completeness, we also report the results of \model/ when $\DL$ is the full training set. Note that in this scenario there is no extra unlabeled data disjoint with the labeled set, and
not surprisingly, \model/ does not outperform the supervised parser.
In addition to the supervised objective~\cref{eq:structVAE:sup_obj} used by the supervised parser, \model/ has the extra unsupervised objective~\cref{eq:structVAE:va_lower_bound}, which uses sampled (probably incorrect) MRs to update the model.
When there is no extra unlabeled data, those sampled (incorrect) MRs add noise to the optimization process, causing \model/ to under-perform.




\begin{figure}[tp]
  \captionsetup{farskip=0pt}%
  \centering
    \subfloat[\django/]{\label{fig:exp:lr:rank:django}{\includegraphics[width=0.5 \columnwidth]{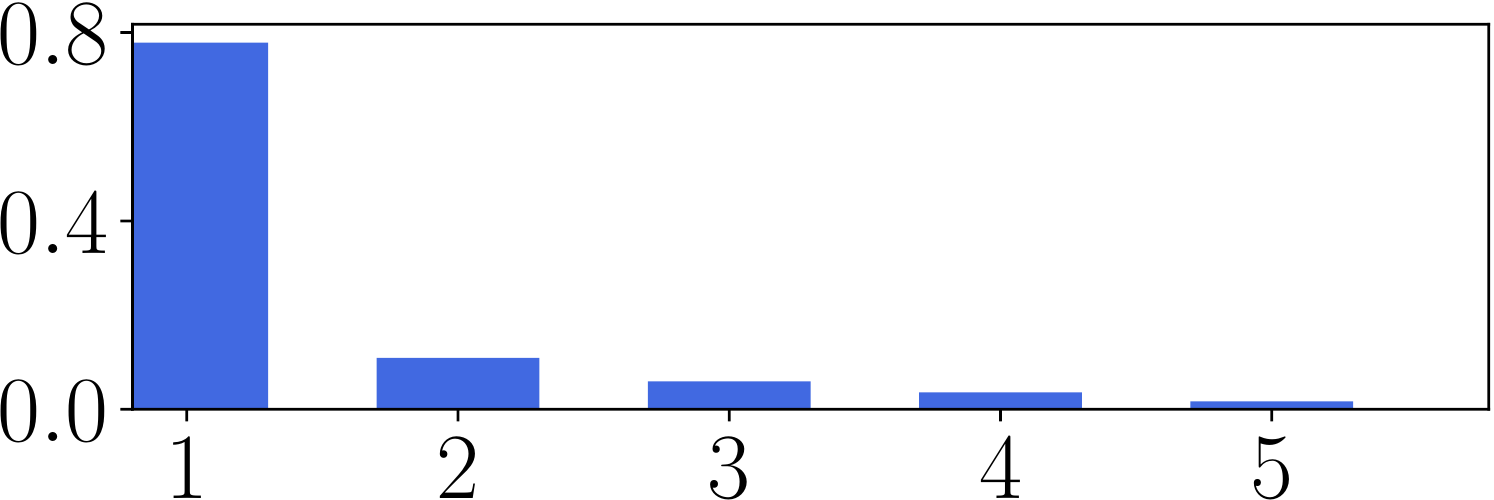} }}%
    \subfloat[\atis/]{\label{fig:exp:lr:rank:atis}{\includegraphics[width=0.5 \columnwidth]{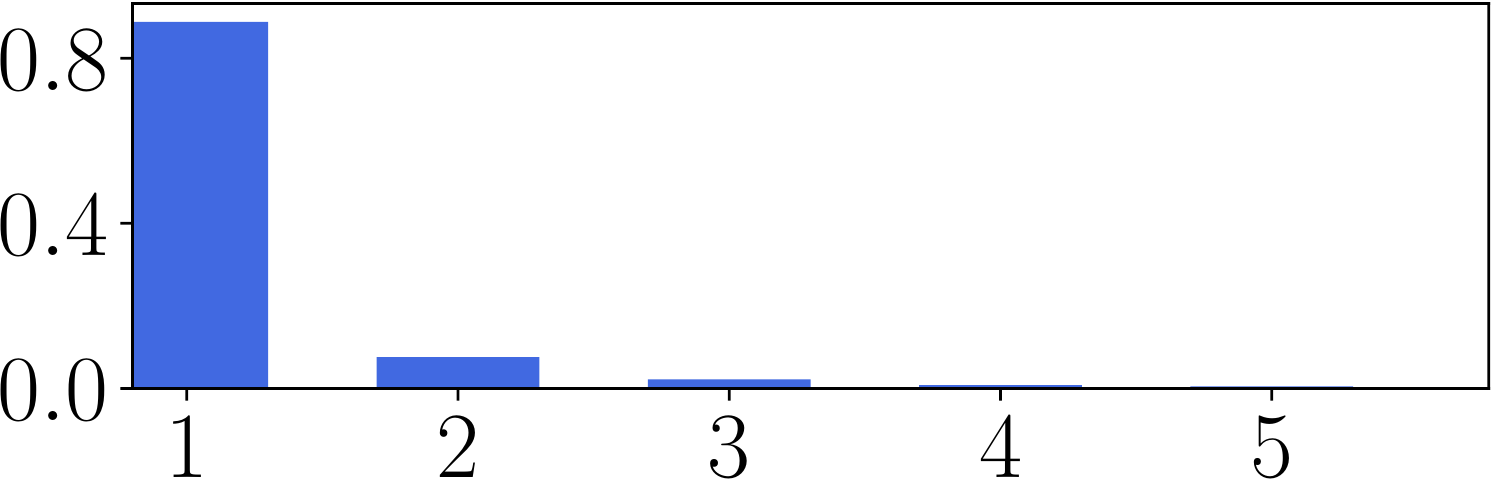} }}%
    \vspace{-2mm}%
    \caption{Distribution of the rank of $l(\x, \mr^*)$ in sampled set}
  \label{fig:exp:lr:rank}
  \vspace{-3mm}
\end{figure}

\paragraph{Study of Learning Signals} As discussed in~\autoref{sec:structVAE:learning}, in semi-supervised learning, the gradient received by the inference model from each sampled latent MR is weighted by the learning signal. 
Empirically, we would expect that on average, the learning signals of gold-standard samples $\mr^*$, $l(\x, \mr^*)$, are positive, larger than those of other (imperfect) samples $\mr'$, $l(\x, \mr')$.
We therefore study the statistics of $l(\x, \mr^*)$ and $l(\x, \mr')$ for all utterances $\x \in \DU - \DL$, \ie the set of utterances which are not included in the labeled set.\footnote{We focus on cases where $\mr^*$ is in the sample set $\mathcal{S}(\x)$.}
The statistics are obtained by performing inference using trained models.
Figures\autoref{fig:exp:lr:hist:django} and\autoref{fig:exp:lr:hist:atis} depict the histograms of learning signals on \django/ and \atis/, resp.
We observe that the learning signals for gold samples concentrate on positive intervals.
We also show the mean and variance of the learning signals.
On average, we have $l(\x, \mr^*)$ being positive and $l(\x, \mr)$ negative.
Also note that the distribution of $l(\x, \mr^*)$ has smaller variance and is more concentrated.
Therefore the inference model receives informative gradient updates to discriminate between gold and imperfect samples.
Next, we plot the distribution of the rank of $l(\x, \mr^*)$, among the learning signals of all samples of $\x$, $\{ l(\x, \mr_i): \mr_i \in \mathcal{S}(\x) \}$.
Results are shown in \autoref{fig:exp:lr:rank}.
We observe that the gold samples $\mr^*$ have the largest learning signals in around 80\% cases. 
We also find that when $\mr^*$ has the largest learning signal, its average difference with the learning signal of the highest-scoring incorrect sample is 1.27 and 0.96 on \django/ and \atis/, respectively.

\begin{table}[tb]
  \centering
  \small
  \setlength\tabcolsep{3pt}
  \begin{tabular}{lp{6.8cm}}
  \toprule
  NL & {\it join p and cmd into a file path, substitute it for f} \\
  {$\mr^s_1$} &
\begin{lstlisting}[basicstyle=\fontfamily{cmtt}\small,style=pythoncode,belowskip=0.2\baselineskip,aboveskip=- 0.5\baselineskip]
f = os.path.join(p, cmd)      |\cmark|
\end{lstlisting} \\
& {$\log q(\mr|\x)=-1.00$ \quad\quad\quad $\log p(\x|\mr)=-2.00$} \\
& {$\log p(\mr)=-24.33$ \quad\quad\quad~ $l(\x, \mr)=9.14$} \\
  {$\mr^s_2$} &
\begin{lstlisting}[basicstyle=\fontfamily{cmtt}\small,style=pythoncode,belowskip=0.2\baselineskip,aboveskip=- 0.5\baselineskip]
p = path.join(p, cmd)      |\xmark|
\end{lstlisting} \\
& {$\log q(\mr|\x)=-8.12$ \quad\quad\quad $\log p(\x|\mr)=-20.96$} \\
& {$\log p(\mr)=-27.89$ \quad\quad\quad~ $l(\x, \mr)=-9.47$} \\
  \hline
  NL & {\it append i-th element of existing to child\_loggers} \\
  {$\mr^s_1$} &
\begin{lstlisting}[basicstyle=\fontfamily{cmtt}\small,style=pythoncode,belowskip=0.2\baselineskip,aboveskip=- 0.5\baselineskip]
child_loggers.append(existing[i])      |\cmark|
\end{lstlisting} \\
& {$\log q(\mr|\x)=-2.38$ \quad\quad\quad $\log p(\x|\mr)=-9.66$} \\
& {$\log p(\mr)=-13.52$ \quad\quad\quad~ $l(\x, \mr)=1.32$} \\
  {$\mr^s_2$} &
\begin{lstlisting}[basicstyle=\fontfamily{cmtt}\small,style=pythoncode,belowskip=0.2\baselineskip,aboveskip=- 0.5\baselineskip]
child_loggers.append(existing[existing])|\xmark|
\end{lstlisting} \\
& {$\log q(\mr|\x)=-1.83$ \quad\quad\quad $\log p(\x|\mr)=-16.11$} \\
& {$\log p(\mr)=-12.43$ \quad\quad\quad~ $l(\x, \mr)=-5.08$} \\
  \hline
  NL & {\it split string pks by ',', substitute the result for primary\_keys} \\
  {$\mr^s_1$} &
\begin{lstlisting}[basicstyle=\fontfamily{cmtt}\small,style=pythoncode,belowskip=0.2\baselineskip,aboveskip=- 0.5\baselineskip]
primary_keys = pks.split(',')      |\cmark|
\end{lstlisting} \\
& {$\log q(\mr|\x)=-2.38$ \quad\quad\quad $\log p(\x|\mr)=-11.39$} \\
& {$\log p(\mr)=-10.24$ \quad\quad\quad~ $l(\x, \mr)=2.05$} \\
  {$\mr^s_2$} &
\begin{lstlisting}[basicstyle=\fontfamily{cmtt}\small,style=pythoncode,belowskip=0.2\baselineskip,aboveskip=- 0.5\baselineskip]
primary_keys = pks.split + ','     |\xmark|
\end{lstlisting} \\
& {$\log q(\mr|\x)=-0.84$ \quad\quad\quad $\log p(\x|\mr)=-14.87$} \\
& {$\log p(\mr)=-20.41$ \quad\quad\quad~ $l(\x, \mr)=-2.60$} \\  
  \bottomrule
  \end{tabular}
  \caption{Inferred latent MRs on \django/ ($|\DL|=5000$). For simplicity we show the surface representation of MRs ($\mr^s$, source code) instead.}
  \label{tab:exp:results:lr:example}
  \vspace{-3.5mm}
\end{table}

Finally, to study the relative contribution of the reconstruction score $\log p(\x|\mr)$ and the prior $\log p(\mr)$ to the learning signal,
we present examples of inferred latent MRs during training (\autoref{tab:exp:results:lr:example}).
Examples 1\&2 show that the reconstruction score serves as an informative quality measure of the latent MR, assigning the correct samples $\mr^s_1$ with high $\log p(\x|\mr)$, leading to positive learning signals.
This is in line with our assumption that a good latent MR should adequately encode the semantics of the utterance.
Example 3 shows that the prior is also effective in identifying ``unnatural'' MRs (\eg~it is rare to add a function and a string literal, as in $\mr^s_2$).
These results also suggest that the prior and the reconstruction model perform well with linearization of MRs.
Finally, note that in Examples 2\&3 the learning signals for the correct samples $\mr^s_1$ are positive even if their inference scores $q(\mr|\x)$ are lower than those of $\mr^s_2$.
This result further demonstrates that learning signals provide informative gradient weights for optimizing the inference model.




\begin{table}[tb]
  \centering
  \resizebox{0.8 \columnwidth}{!}{
  \begin{tabular}{lcc}
  \toprule
  \textbf{$|\DL|$} & \textsc{Supervised} & {\model/-\textsc{Seq}} \\
  \hline
    500 & 47.3 & {\bf 55.6} \\
    1,000 & 62.5 & {\bf 73.1} \\
    2,000 & 73.9 & {\bf 74.8} \\
    3,000 & 80.6 & {\bf 81.3} \\ \hdashline
    4,434 (All) & {\bf 84.6} & 84.2 \\
  \bottomrule \end{tabular}}
  \caption{Performance of the \model/-{\sc Seq} on \atis/ w.r.t.~the size of labeled training data $\DL$}
  \label{tab:exp:results:seq}
  \vspace{-3mm}
\end{table}

\paragraph{Generalizing to Other Latent MRs}
Our main results are obtained using a strong AST-based semantic parser as the inference model, with copy-augmented reconstruction model and an LSTM language model as the prior. 
However, there are many other ways to represent and infer structure in semantic parsing~\citep{Carpenter:1998:TS:286695,Steedman:2000:SP:332037}, and thus it is of interest whether our basic \model/ framework generalizes to other semantic representations.
To examine this, we test \model/ using $\lambda$-calculus logical forms as latent MRs for semantic parsing on the \atis/ domain.
We use standard sequence-to-sequence networks with attention~\citep{luong2015effective} as inference and reconstruction models.
The inference model is trained to construct a tree-structured logical form using the transition actions defined in~\citet{cheng2017learning}.
We use a classical tri-gram Kneser-Ney language model as the prior.
\autoref{tab:exp:results:seq} lists the results for this \model/-{\sc Seq} model.

We can see that even with this very different model structure \model/ still provides significant gains, demonstrating its compatibility with different inference/reconstruction networks and priors.
Interestingly, compared with the results in~\autoref{tab:exp:results:atis}, we found that the gains are especially larger with few labeled examples --- \model/-{\sc Seq} achieves improvements of 8-10 points when $|\DL|<1000$.
These results suggest that semi-supervision is especially useful in improving a mediocre parser in low resource settings. 


\begin{table}[tb]
  \small
  \centering
  \resizebox{\columnwidth}{!}{%
  \begin{tabular}{llll|llll}
  \hline

  \hline
  \multicolumn{4}{c|}{\atis/} & \multicolumn{4}{c}{\django/} \\
  $|\DL|$ & \textsc{Sup.} & \textsc{mlp} & \textsc{lm} & $|\DL|$ & \textsc{Sup.} & \textsc{mlp} & \textsc{lm} \\
  \hline
  500  & 63.2 & {\it 61.5$^\dagger$} & {\bf 66.0} & 1,000 & 49.9 & {\it 47.0$^\dagger$} & {\bf 52.0} \\
  1,000 & 74.6 & {\bf 76.3} & 75.7 & 5,000 & 63.2 & {\it 62.5$^\dagger$} & {\bf 65.6} \\
  2,000 & 80.4 & {\bf 82.9} & 82.4 & 8,000 & 70.3 & {\it 67.6$^\dagger$} & {\bf 71.5} \\
  3,000 & 82.8 & {\it 81.4$^\dagger$} & {\bf 83.6} & 12,000 & 71.1 & 71.6 & {\bf 72.0} \\
  \hline

  \hline
  \end{tabular}}
  \caption{Comparison of \model/ with different baseline functions $b(\x)$, {\it italic$^\dagger$}: semi-supervised learning with the MLP baseline is worse than supervised results.}
  \label{tab:exp:baseline}
  \vspace{-3.5mm}
\end{table}

\paragraph{Impact of Baseline Functions} In \autoref{sec:structVAE:learning} we discussed our design of the baseline function $b(\x)$ incorporated in the learning signal (\cref{eq:framework:semi_sup:encoder_grad}) to stabilize learning, which is based on a language model (LM) over utterances (\cref{eq:structVAE:baseline}).
We compare this baseline with a commonly used one in REINFORCE training: the multi-layer perceptron (MLP). The MLP takes as input the last hidden state of the utterance given by the encoding LSTM of the inference model.
\autoref{tab:exp:baseline} lists the results over sampled settings.
We found that although \model/ with the MLP baseline sometimes registers better performance on \atis/, in most settings it is worse than our LM baseline, and could be even worse than the supervised parser.
On the other hand, our LM baseline correlates well with the learning signal, yielding stable improvements over the supervised parser.
This suggests the importance of using carefully designed baselines in REINFORCE learning, especially when the reward signal has large range (\eg log-likelihoods).




\begin{figure}[tb]
  \centering
  \includegraphics[width=0.8 \columnwidth]{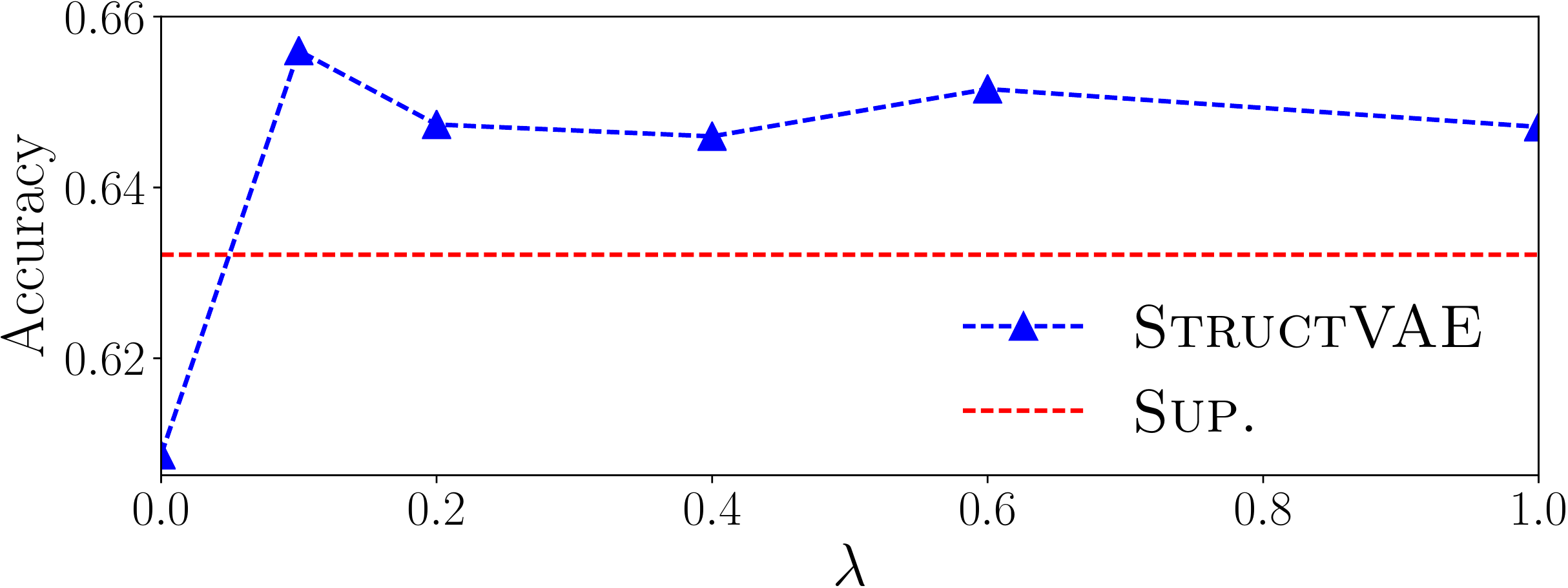}
  \vspace{-2mm}
  \caption{Performance on \django/ ($|\DL|=5000$) w.r.t.~the KL weight $\lambda$}
  \label{fig:exp:results:kl}
  \vspace{-3mm}
\end{figure}

\begin{figure}[tb]
  \centering
  \includegraphics[width=0.8 \columnwidth]{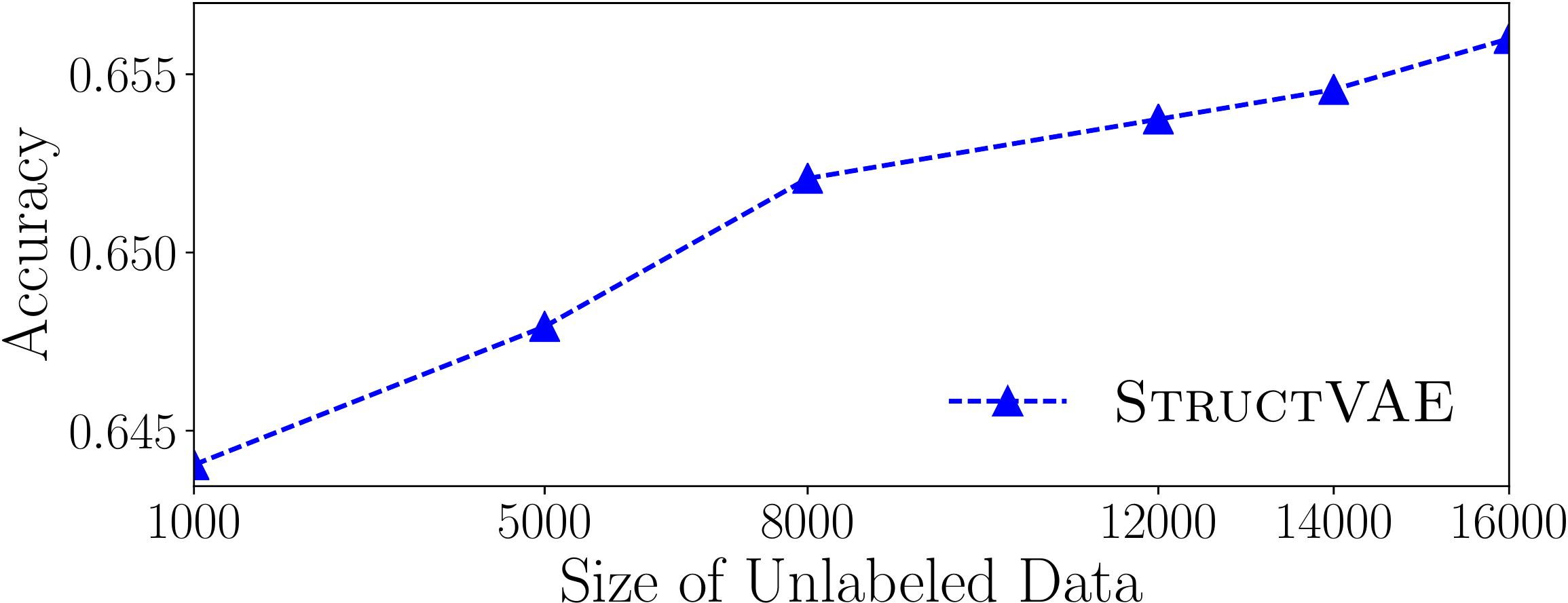}
  \vspace{-1mm}
  \caption{Performance on \django/ ($|\DL|=5000$) w.r.t.~the size of unlabeled data $\DU$}
  \label{fig:exp:results:unlabeled_size}
  \vspace{-3mm}
\end{figure}

\paragraph{Impact of the Prior $p(\mr)$}
\autoref{fig:exp:results:kl} depicts the performance of \model/ as a function of the KL term weight $\lambda$ in~\cref{eq:structVAE:va_lower_bound}.
When \model/ degenerates to a vanilla auto-encoder without the prior distribution (\ie $\lambda=0$), it under-performs the supervised baseline. 
This is in line with our observation in \autoref{tab:exp:results:lr:example} showing that the prior helps identify unnatural samples.
The performance of the model also drops when $\lambda > 0.1$, suggesting that empirically controlling the influence of the prior to the inference model is important.

\paragraph{Impact of Unlabeled Data Size} \autoref{fig:exp:results:unlabeled_size} illustrates the accuracies w.r.t.~the size of unlabeled data.
\model/ yields consistent gains as the size of the unlabeled data increases.

\vspace{-0.5mm}
\section{Related Works}
\vspace{-0.5mm}

\paragraph{Semi-supervised Learning for NLP} 
Semi-supervised learning comes with a long history~\citep{zhu05survey}, with applications in NLP from early work of self-training~\citep{yaro95word}, and graph-based methods~\citep{das11frame}, to recent advances in auto-encoders~\citep{cheng16seminmt,socher11semisentiment,zhang17semicrf} and deep generative methods~\citep{xu17semitextclassification}.
Our work follows the line of neural variational inference for text processing~\citep{miao2016variational}, and resembles~\citet{miao2016language}, which uses VAEs to model summaries as discrete latent variables for semi-supervised summarization, while we extend the VAE architecture for more complex, tree-structured latent variables.

\paragraph{Semantic Parsing}
Most existing works alleviate issues of limited parallel data through weakly-supervised learning, using the denotations of MRs as indirect supervision~\citep{reddy_largescale_2014,DBLP:conf/emnlp/KrishnamurthyTK16,DBLP:journals/corr/NeelakantanLS15,pasupat2015compositional,DBLP:dblp_conf/ijcai/YinLLK16}. 
For semi-supervised learning of semantic parsing, \citet{kate07semisp} first explore using transductive SVMs to learn from a semantic parser's predictions.
\citet{konstas2017neural} apply self-training to bootstrap an existing parser for AMR parsing.
\citet{Kocisky2016} employ VAEs for semantic parsing, but in contrast to \model/'s structured representation of MRs, they model NL utterances as flat latent variables, and learn from unlabeled MR data.
There have also been efforts in unsupervised semantic parsing, which exploits external linguistic analysis of utterances (\eg dependency trees) and the schema of target knowledge bases to infer the latent MRs~\citep{poon09usp,poon13grounded}.
Another line of research is domain adaptation, which seeks to transfer a semantic parser learned from a source domain to the target domain of interest, therefore alleviating the need of parallel data from the target domain~\citep{su17crossdomain,xing17transfer,herzig18zeroshot}.

\section{Conclusion}
We propose \model/, a deep generative model with tree-structured latent variables for semi-supervised semantic parsing.
We apply \model/ to semantic parsing and code generation tasks, and show it outperforms a strong supervised parser using extra unlabeled data.


\bibliography{structVAE}
\bibliographystyle{acl_natbib}

\newpage
\clearpage
\appendix
\onecolumn
\begin{center}
\Large
\textbf{\model/: Tree-structured Latent Variable Models for Semi-supervised Semantic Parsing}

Supplementary Materials
\end{center}
\section{Generating Samples from \model/}
\model/ is a generative model of natural language, and therefore can be used to sample latent MRs and the corresponding NL utterances. This amounts to draw a latent MR $\mr$ from the prior $p(\mr)$, and sample an NL utterance $\x$ from the reconstruction model $p_{\theta}(\x | \mr)$. 
Since we use the sequential representation $\mr^s$ in the prior, to guarantee the syntactic well-formedness of sampled MRs from $p(\mr)$, we use a syntactic checker and reject any syntactically-incorrect samples\footnote{We found most samples from $p(\mr)$ are syntactically well-formed, with 98.9\% and 95.3\% well-formed samples out of 100$K$ samples on \atis/ and \django/, respectively.}. 
\autoref{tab:app:sample:django} and \autoref{tab:app:sample:atis} present samples from \django/ and \atis/, respectively.
These examples demonstrate that \model/ is capable of generating syntactically diverse NL utterances.

\begin{table}[h]
	\centering
	\small
	\begin{tabular}{rl}
	\toprule
	latent MR &
\begin{lstlisting}[basicstyle=\fontfamily{cmtt}\small,style=pythoncode]
def __init__(self, *args, **kwargs): pass
\end{lstlisting} \\
	surface NL & {\it Define the method \_\_init\_\_ with 3 arguments: self, unpacked list args and unpacked dictionary kwargs} \\
	\hline
	latent MR &
\begin{lstlisting}[basicstyle=\fontfamily{cmtt}\small,style=pythoncode]
elif isinstance(target, six.string_types): pass
\end{lstlisting} \\
	surface NL & {\it Otherwise if target is an instance of six.string\_types} \\
	\hline
	latent MR &
\begin{lstlisting}[basicstyle=\fontfamily{cmtt}\small,style=pythoncode]
for k, v in unk.items(): pass
\end{lstlisting} \\
  surface NL & {\it For every k and v in return value of the method unk.items} \\
  \hline
  latent MR &
\begin{lstlisting}[basicstyle=\fontfamily{cmtt}\small,style=pythoncode]
return cursor.fetchone()[0]
\end{lstlisting} \\
	surface NL & {\it Call the method cursor.fetchone , return the first element of the result} \\
	\hline
	latent MR &
\begin{lstlisting}[basicstyle=\fontfamily{cmtt}\small,style=pythoncode]
sys.stderr.write(_STR_ % e)
\end{lstlisting} \\
	surface NL & {\it Call the method sys.stderr, write with an argument \_STR\_ formated with e} \\
	\hline
	latent MR &
\begin{lstlisting}[basicstyle=\fontfamily{cmtt}\small,style=pythoncode]
opts = getattr(self, _STR_, None)
\end{lstlisting} \\
	surface NL & {\it Get the \_STR\_ attribute of the self object, if it exists substitute it for opts, if not opts is None} \\
	\bottomrule
	\end{tabular}
	\caption{Sampled latent meaning representations (presented in surface source code) and NL utterances from \django/.}
	\label{tab:app:sample:django}
\end{table}

\begin{table}[h]
	\centering
	\small
	\begin{tabular}{rl}
	\toprule
	latent MR &
\begin{lstlisting}[basicstyle=\fontfamily{cmtt}\small,style=atiscode,belowskip=-\baselineskip,aboveskip=- 0.6\baselineskip]
(argmax $0 (and (flight $0) (meal $0 lunch:me) 
                  (from $0 ci0) (to $0 ci1)) (departure_time $0))
\end{lstlisting} \\
	surface NL & {\it Show me the latest flight from ci0 to ci1 that serves lunch} \\
	\hline
	latent MR &
\begin{lstlisting}[basicstyle=\fontfamily{cmtt}\small,style=atiscode,belowskip=-\baselineskip,aboveskip=- 0.6\baselineskip]
(min $0 (exists $1 (and (from $1 ci0) (to $1 ci1) (day_number $1 dn0) 
                           (month $1 mn0) (round_trip $1) (= (fare $1) $0))))
\end{lstlisting} \\
	surface NL & {\it I want the cheapest round trip fare from ci0 to ci1 on mn0 dn0} \\
	\hline
	latent MR &
\begin{lstlisting}[basicstyle=\fontfamily{cmtt}\small,style=atiscode,belowskip=-\baselineskip,aboveskip=- 0.6\baselineskip]
(lambda $0 e (and (flight $0) (from $0 ci0) (to $0 ci1) (weekday $0)))
\end{lstlisting} \\
	surface NL & {\it Please list weekday flight between ci0 and ci1} \\
	\hline
	latent MR &
\begin{lstlisting}[basicstyle=\fontfamily{cmtt}\small,style=atiscode,belowskip=-\baselineskip,aboveskip=- 0.6\baselineskip]
(lambda $0 e (and (flight $0) (has_meal $0) (during_day $0 evening:pd) 
		    (from $0 ci1) (to $0 ci0) (day_number $0 dn0) (month $0 mn0)))
\end{lstlisting} \\
	surface NL & {\it What are the flight from ci1 to ci0 on the evening of mn0 dn0 that serves a meal} \\
	\hline
  latent MR &
\begin{lstlisting}[basicstyle=\fontfamily{cmtt}\small,style=atiscode,belowskip=-\baselineskip,aboveskip=- 0.6\baselineskip]
(lambda $0 e (and (flight $0) (oneway $0) (class_type $0 first:cl) (from $0 ci0)
                    (to $0 ci1) (day $0 da0)))
\end{lstlisting} \\
  surface NL & {\it Show me one way flight from ci0 to ci1 on a da0 with first class fare} \\
  \hline
	latent MR &
\begin{lstlisting}[basicstyle=\fontfamily{cmtt}\small,style=atiscode,belowskip=-\baselineskip,aboveskip=- 0.6\baselineskip]
(lambda $0 e (exists $1 (and (rental_car $1) (to_city $1 ci0) 
                                (= (ground_fare $1) $0))))
\end{lstlisting} \\
	surface NL & {\it What would be cost of car rental car in ci0} \\
	\bottomrule
	\end{tabular}
	\caption{Sampled latent meaning representations (presented in surface $\lambda$-calculus expression) and NL utterances from \atis/. Verbs are recovered to their correct form instead of the lemmatized version as in the pre-processed dataset.}
	\label{tab:app:sample:atis}
\end{table}

\section{Neural Network Architecture}
\label{sec:app:neuralnet}

\subsection{Prior $p(\mr)$}
\label{sec:app:neuralnet:prior}

The prior $p(\mr)$ is a standard LSTM language model~\citep{zaremba14lstm}.
We use the sequence representation of $\mr$, $\mr^s$, to model $p(\mr)$. Specifically, let $\mr^s = \{ z^s_i \}_{i = 1}^{|\mr^s|}$ consisting of $|\mr^s|$ tokens, we have
\begin{equation*}
  p(\mr^s) = \prod_{i = 1}^{|\mr^s|} p(z^s_i | \mr^s_{<i}),
\end{equation*}
where $\mr^s_{<i}$ denote the sequence of history tokens $\{ z^s_1, z^s_2, \ldots, z^s_{i - 1} \}$. At each time step $i$, the probability of predicting $z^s_i$ given the context is modeled by an LSTM network
\begin{gather*}
  p(z^s_i | \mr^s_{<i}) = \textrm{softmax}( \bm{W}\bm{h}_i + \bm{b}) \\
  \bm{h}_i = f_{\textrm{LSTM}}(\bm{e}(z^s_{i-1}), \bm{h}_{i-1})
\end{gather*}
where $\bm{h}_i$ denote the hidden state of the LSTM at time step $i$, and $\bm{e}(\cdot)$ is an embedding function.

\subsection{Reconstruction Model $p_{\theta}(\x|\mr)$}
\label{sec:app:neuralnet:reconstruction_model}

We implement a standard attentional sequence-to-sequence network~\citep{luong2015effective} with copy mechanism as the reconstruction network $p_{\theta}(\x|\mr)$.
Formally, given a utterance $\x$ of $n$ words $\{ x_i \}_{i=1}^n$,
the probability of generating a token $x_i$ is marginalized over the probability of generating $x_i$ from a closed-set vocabulary, and that of copying from the MR $\mr^s$:
\begin{align*}
  p(x_i | x_{<i}, \mr^s) &=  p(\textrm{gen}|x_{<i}, \mr^s) p(x_i|\textrm{gen}, x_{<i}, \mr^s) \\ &\quad\quad + p(\textrm{copy}|x_{<i}, \mr^s) p(x_i|\textrm{copy}, x_{<i}, \mr^s)
\end{align*}
where $p(\textrm{gen}|\cdot)$ and $p(\textrm{copy}|\cdot)$ are computed by $\textrm{softmax}(\bm{W} \bm{\tilde{s}}_i^c)$. $\bm{\tilde{s}}_i^c$ denotes the attentional vector~\citep{luong2015effective} at the $i$-th time step:
\begin{equation}
  \bm{\tilde{s}}_i^c = \tanh(\bm{W}_c[\bm{c}^c_i; \bm{s}^c_i]).
  \label{eq:att_vec}
\end{equation}
Here, $\bm{s}^c_i$ is the $i$-th decoder hidden state of the reconstruction model, and $\bm{c}^c_i$ the context vector~\citep{Bahdanau2015} obtained by attending to the source encodings.
The probability of copying the $j$-th token in $\mr^s$, $z^s_j$, is given by a pointer network~\citep{DBLP:conf/nips/VinyalsFJ15}, derived from $\bm{\tilde{s}}_i^c$ and the encoding of $z^s_j$, $\bm{h}_j^{\mr}$.
\begin{equation*}
  p(x_i = z^s_j|\textrm{copy}, x_{<i}, \mr^s) = \frac{ \exp ( {\bm{h}_j^{\mr}}^\intercal \bm{W} \bm{\tilde{s}}_i^c ) } {\sum_{j'=1}^{|\mr^s|} \exp ( {\bm{h}_{j'}^{\mr}}^\intercal \bm{W} \bm{\tilde{s}}_i^c ) }
\end{equation*}

\subsection{Inference Model $p_{\phi}(\mr|\x)$}
\label{sec:app:neuralnet:inference_model}
Our inference model (\ie the semantic parser) is based on the code generation model proposed in~\citet{yin17acl}. As illustrated in~\autoref{fig:ast_gen_example} and elaborated in~\autoref{sec:inference}, our transition parser constructs an abstract syntax tree specified under the ASDL formalism using a sequence of transition actions.
The parser is a neural sequence-to-sequence network, whose recurrent decoder is augmented with auxiliary connections following the topology of ASTs. 
Specifically, at each decoding time step $t$, an LSTM decoder uses its internal hidden state $\bm{s}_t$ to keep track of the generation process of a derivation AST
\begin{equation*}
  \bm{s}_t = f_{\textrm{LSTM}}([\bm{a}_{t-1}: \bm{\tilde{s}}_{t-1}: \bm{p}_t], \bm{s}_{t-1})
\end{equation*}
where $[:]$ denotes vector concatenation. 
$\bm{a}_{t-1}$ is the embedding of the previous action.
$\bm{\tilde{s}}_{t-1}$ is the input-feeding attentional vector as in~\citet{luong2015effective}.
$\bm{p}_t$ is a vector that captures the information of the parent frontier field in the derivation AST, which is the concatenation of four components:
$\bm{n_{f_t}}$, which is the embedding of the current frontier field $n_{f_t}$ on the derivation;
$\bm{e_{f_t}}$, which is the embedding of the type of $n_{f_t}$;
$\bm{s}_{p_t}$, which is the state of the decoder at which the frontier field $n_{f_t}$ was generated by applying its parent constructor $c_{p_t}$ to the derivation;
$\bm{c}_{p_t}$, which is the embedding of the parent constructor $c_{p_t}$.

Given the current state of the decoder, $\bm{s}_t$, an attentional vector $\bm{\tilde{s}}_t$ is computed similar as~\cref{eq:att_vec} by attending to input the utterance $\x$.
The attentional vector $\bm{\tilde{s}}_t$ is then used as the query vector to compute action probabilities, as elaborated in \S4.2.2 of~\citet{yin17acl}.

\section{ASDL Grammar for \atis/}
\label{sec:app:asdl_atis}

We use the ASDL grammar defined in~\citet{rabinovich17syntaxnet} to deterministically convert between $\lambda$-calculus logical forms and ASDL ASTs:

\begin{lstlisting}[basicstyle=\fontfamily{cmtt}\small,xleftmargin=.15\textwidth]
expr = Variable(var variable)
     | Entity(ent entity)
     | Number(num number)
     | Apply(pred predicate, expr* arguments)
     | Argmax(var variable, expr domain, expr body)
     | Argmin(var variable, expr domain, expr body)
     | Count(var variable, expr body)
     | Exists(var variable, expr body)
     | Lambda(var variable, var_type type, expr body)
     | Max(var variable, expr body)
     | Min(var variable, expr body)
     | Sum(var variable, expr domain, expr body)
     | The(var variable, expr body)
     | Not(expr argument)
     | And(expr* arguments)
     | Or(expr* arguments)
     | Compare(cmp_op op, expr left, expr right)

cmp_op = Equal | LessThan | GreaterThan
\end{lstlisting}

\section{Model Configuration}
\label{sec:app:config}


\paragraph{Initialize Baselines $b(\x)$} 
\model/ uses baselines $b(\x)$ to reduce variance in training.
For our proposed baseline based on the language model over utterances (\cref{eq:structVAE:baseline}), we pre-train a language model using all NL utterances in the datasets.
For terms $a$ and $c$ in~\cref{eq:structVAE:baseline}, we determine their initial values by first train \model/ starting from $a=1.0$ and $c=0$ for a few epochs, and use their optimized values. Finally we initialize $a$ to 0.5 and $b$ to $-2.0$ for \atis/, and $a$ to 0.9 and $b$ to 2.0 for \django/.
We perform the same procedure to initialize the bias term $b_\textrm{MLP}$ in the MLP baseline, and have $b_\textrm{MLP} = -20.0$.

\paragraph{Pre-trained Priors $p(\mr)$} 
\model/ requires pre-trained priors $p(\mr)$~(\autoref{sec:structVAE:learning}).
On \atis/, we train a prior for each labeled set $\DL$ of size $K$ using the MRs in $\DL$.
For \django/, we use all source code in Django that is not included in the annotated dataset.

\paragraph{Hyper-Parameters and Optimization} For all experiments we use embeddings of size 128, and LSTM hidden size of 256.
For the transition parser, we use the same hyper parameters as~\citet{yin17acl}, except for the node (field) type embedding, which is 64 for \django/ and 32 for \atis/.
To avoid over-fitting, we impose dropouts on the LSTM hidden states, with dropout rates validated among $\{ 0.2, 0.3, 0.4 \}$.
We train the model using Adam~\citep{kingma14adam}, with a batch size of 10 and 25 for the supervised and unsupervised objectives, resp.
We apply early stopping, and reload the best model and halve the learning rate when the performance on the development set does not increase after 5 epochs. We repeat this procedure for 5 times.

\section{\textsc{\textbf{Seq2Tree}} Results on \atis/ Data Splits}

\begin{table}[H]
  \vspace{-3mm}
  \centering
  \small
  \begin{tabular}{l:cc}
  \hline

  \hline
  \textbf{$|\DL|$} & \textsc{Sup.} & \sq/ \\
  \hline
    500 & 63.2 & 57.1  \\
    1000 & 74.6 & 69.9 \\
    2000 & 80.4 & 71.7 \\
    3000 & 82.8 & 81.5 \\
  \hline

  \hline
  \end{tabular}
  \caption{Accuracies of \sq/ and our supervised parser on different data splits of \atis/}
  \label{tab:app:sq}
  \vspace{-3mm}
\end{table}

We also present results of \sq/~\citep{DBLP:conf/acl/DongL16} trained on the data splits used in~\autoref{tab:exp:results:atis}, as shown in~\autoref{tab:app:sq}. 
Our supervised parser performs consistently better than~\sq/.
This is probably due to the fact that our transition-based parser encodes the grammar of the target logical form {\it a priori} under the ASDL specification, in contrast with \sq/ which need to learn the grammar from the data. This would lead to improved performance when the amount of parallel training data is limited.

\end{document}